\documentclass[10pt,twocolumn,letterpaper]{article}

\usepackage[pagenumbers]{wacv} 
\usepackage{graphicx}
\usepackage{amsmath}
\usepackage{amssymb}
\usepackage{booktabs}

\usepackage[pagebackref,breaklinks,colorlinks]{hyperref}

\usepackage[capitalize]{cleveref}
\crefname{section}{Sec.}{Secs.}
\Crefname{section}{Section}{Sections}
\Crefname{table}{Table}{Tables}
\crefname{table}{Tab.}{Tabs.}

\def\wacvPaperID{1088} 
\def\confName{WACV}
\def\confYear{2025}

\usepackage{xcolor}

\usepackage{latexsym}
\usepackage{amssymb}
\usepackage{amsmath}
\usepackage{booktabs}
\usepackage{enumitem}
\usepackage{graphicx}
\usepackage{color}
\usepackage{multirow}
\usepackage{algorithm}
\usepackage{algorithmic}
\usepackage{tcolorbox}

\begin{document}

\title{Adaptive Deviation Learning for Visual Anomaly Detection with Data Contamination}


\author{
    Anindya Sundar Das\textsuperscript{1},
    Guansong Pang\textsuperscript{2},
    Monowar Bhuyan\textsuperscript{1} \\
    \textsuperscript{1}Ume{\aa} University, Sweden \\
    \textsuperscript{2}Singapore Management University, Singapore\\
    \textsuperscript{1}{\tt\small \{aninsdas, monowar\}@cs.umu.se},
    \textsuperscript{2}{\tt\small gspang@smu.edu.sg}
}

\maketitle

\begin{abstract}
   Visual anomaly detection targets to detect images that notably differ from normal pattern, and it has found extensive application in identifying defective parts within the manufacturing industry. These anomaly detection paradigms predominantly focus on training detection models using only clean, unlabeled normal samples, assuming an absence of contamination; a condition often unmet in real-world scenarios. The performance of these methods significantly depends on the quality of the data and usually decreases when exposed to noise. We introduce a systematic adaptive method that employs deviation learning to compute anomaly scores end-to-end while addressing data contamination by assigning relative importance to the weights of individual instances. In this approach, the anomaly scores for normal instances are designed to approximate scalar scores obtained from the known prior distribution. Meanwhile, anomaly scores for anomaly examples are adjusted to exhibit statistically significant deviations from these reference scores. Our approach incorporates a constrained optimization problem within the deviation learning framework to update instance weights, resolving this problem for each mini-batch. Comprehensive experiments on the MVTec and VisA benchmark datasets indicate that our proposed method surpasses competing techniques and exhibits both stability and robustness in the presence of data contamination.\footnote[1]{Our code is available at https://github.com/anindyasdas/ADL4VAD/}
\end{abstract}

\section{Introduction}
\label{sec:intro}
Visual anomaly detection (VAD) seeks to identify unusual patterns within established normal visual data that are specific to a particular domain. It is of great interest because of its diverse applications, including industrial defect detection \cite{roth2022towards, zavrtanik2021draem}, medical lesion detection \cite{schlegl2019f}, video surveillance \cite{liu2018classifier}, and many others. Most existing solutions train learning models using a substantial amount of clean, normal data, assuming such data is readily available in real-world scenarios. However, in numerous real-world applications, the data often obtained is contaminated or contains noisy labels due to errors in the human annotation process or labels inherited from legacy systems. This can lead to abnormal behavior when such systems are deployed in real industrial settings. While there has been extensive research addressing learning under the assumption of noisy labels, very few studies specifically focused on visual anomaly detection in corrupted data settings.

Prior efforts to address the generic problem of learning with noisy labels include identifying and correcting label noise \cite{hendrycks2018using}, label smoothing \cite{szegedy2016rethinking}, and devising robust loss functions \cite{ma2020normalized}, etc.. In a recent work \cite{kumar2021constrained}, instance reweighting is employed to dynamically assign relative importance weights to data samples. This technique aims to reduce the impact of potentially noisy examples, as noisy or mislabeled instances tend to have higher loss values \cite{arazo2019unsupervised, majidi2021exponentiated} compared to clean ones.

Anomaly Detection (AD) methods predominantly focus on normal samples to train networks. However, some recent works also utilize both normal and anomalous clean samples for training. Since obtaining large-scale anomalous data with accurate labels is both expensive and challenging, these methods utilize a limited number of labeled anomalous samples to train the network. Deviation Learning \cite{pang2021explainable, pang2019deep, das2023few} leverages a small set of labeled anomalous examples to train an anomaly-conscious detection networks. In this work, we adapted deviation learning within a data contamination setup to learn an anomaly score in a self-supervised manner, thereby introducing a novel approach that enhances model robustness.

\textit{The primary contributions of our work can be summarized as follows: i) We propose a novel adaptive deviation learning framework that integrates constrained optimization into our self-supervised learning setup to dynamically assign sample importance weights to our deviation loss-based objective, thereby enhancing the robustness of anomaly score learning. ii) We also introduce a soft deviation-based objective that utilizes the likelihood of each instance being an anomaly instead of relying on hard labels, which effectively addresses corrupt data. iii) We perform comprehensive experiments to set performance benchmarks for our proposed method on two publicly available datasets. iv) We provide a detailed analysis of various state-of-the-art methods, examining how increased contamination affects their performance across different metrics. v) Our study demonstrates that our method attains state-of-the-art performance in anomaly detection under data contamination setup, supported by extensive empirical results from the MVTec and VisA datasets.}

\section{Related Work}


\noindent \textbf{Anomaly Detection on Clean Data.} Classical methods typically concentrate on modeling a one-class distribution using only normal samples. These methods \cite{ruff2018deep, tax2004support, yi2020patch} attempt to find decision boundaries around normal data  to distinguish it from anomalous samples, by assuming that the normal data follows a unimodal distribution. 
In \textbf{reconstruction-based} methods, autoencoders \cite{bergmann2018improving}, generative adversarial networks (GANs) \cite{akccay2019skip, tang2020anomaly, perera2019ocgan, schlegl2019f, schlegl2017unsupervised}, variational autoencoders (VAEs) \cite{vasilev2020q} are usually trained on clean normal data and the anomaly score is devised as the difference between input images and reconstructed images at the output. In DR\AE M \cite{zavrtanik2021draem}, the generative network is combined with the discriminative network, which enhances overall
performance. Nevertheless, these models' effectiveness could still suffer from problems stemming from over-generalization \cite{pirnay2022inpainting}. A \textbf{pre-trained} network, trained on large-scale datasets, can also serve for anomaly detection \cite{defard2021padim, deng2022anomaly, rippel2021modeling}. 
Another distance-based approach \cite{defard2021padim, rippel2021modeling, pang2021explainable} involves utilizing a \textbf{prior distribution}, such as a Gaussian distribution, to model the feature distribution of normal samples. Recent researches in \textbf{few-shot learning} \cite{pang2021explainable, pang2019deep} highlight weakness in one-class classification and leverage a
small set of labeled anomalous examples, thereby integrating prior
knowledge of anomalies into the model. These models uses deviation loss\cite{pang2021explainable} based objective, and anomaly score is computed as deviation from the mean of some prior reference score drawn from a known distribution. In our approach, we utilize a deviation loss approach within a data contamination scenario to learn an anomaly score in a self-supervised manner. In \textbf{memory-based} approaches \cite{yi2020patch, roth2022towards}, a memory bank of normal data is constructed during training. Inference involves selecting the nearest item in the memory bank when presented with a query item. The similarity between the query item and the nearest item is then utilized to calculate the anomaly score. In \textbf{data augmentation-based} methods the standard one-class classification AD is  converted into supervised task by simulating pseudo-anomalies. As the classical augmentation techniques \cite{gidaris2018unsupervised, devries2017improved} do not perform well in detecting fine-grained anomalies, new strategies \cite{li2021cutpaste, yun2019cutmix} have been developed. These strategies involve randomly selecting a rectangular region from a source image and copying and pasting the content to a different location  within the target image. Some other approaches \cite{zavrtanik2021draem, zavrtanik2022dsr, zhang2023destseg} use perlin noise \cite{perlin1985image} to simulate more-realistic anomalous images. In our proposed approached we use perlin noise to simulate pseudo-anomalies for self-supervision task.

\noindent \textbf{Anomaly Detection on Contaminated Data.} In many real-world scenarios, training data is frequently contaminated with unlabelled anomalies resulting in noisy dataset. Training AD models on these noisy datasets  significantly undermines the models' performance \cite{wang2019effective, huyan2022unsupervised}. In the 'Blind' training strategy, the contaminated dataset is treated as if it were clean, with methods relying on prioritizing inliers \cite{wang2019effective}. These techniques work well when contamination levels are low. Alternatively, other strategies \cite{yoon2021self} involve iteratively removing anomalies from contaminated data using an ensemble of classifiers; then the anomaly detector is trained on refined dataset. However, these methods often fall short in leveraging the insights gained from outliers. Outlier Exposure methods \cite{hendrycks2018deep, qiu2022latent}, on the contrary, utilize additional training signals from anomalies present in the corrupted dataset.

\section{ADL: Adaptive Deviation Learning}
In this section, we begin by explaining the problem statement, followed by introduction to our proposed framework, including its constituent components, and the underlying mathematical principles. Finally, we explain our proposed learning algorithm, encompassing the associated losses and the optimization procedure.

\subsection{Problem Formulation}

In this paper, our objective is to explicitly learn anomaly scores using a self-supervised approach, where the unlabeled training dataset, primarily composed of normal data, is corrupted with instances of anomalies. The training dataset $\mathcal{X_N} = \{ x_1, x_2, ..., x_{i}, ..., x_{N} \} $ consisting of $N$ data samples. A fraction $\epsilon$ of the data is outliers or anomalies, while the remaining $(1-\epsilon)$ fraction represents normal instances or inliers. For each instance $x_i \in \mathcal{X_N}$, we designate $y_i=0$ if the instance is normal and $y_i=1$ if it is anomalous. Notably, our framework operates without presumptions or prior knowledge regarding the true contamination ratio $\epsilon$, and assume the data $\mathcal{X_N}$ to be clean. To train the model discriminatively, we synthetically generate a set of pseudo-anomalies, denoted as $\mathcal{X_P} = \{ x_1, x_2, ..., x_{j}, ..., x_{M} \}$, where $y_j=1$ if $x_j \in \mathcal{X_P}$. Our goal to learn an anomaly scoring function, denoted as $\psi_K: \mathcal{X} \rightarrow \mathbb{R}$ which assign scores to data samples indicating whether they are anomalous or normal. The aim is to ensure that $\psi_K(x_i)$ is close to the mean of anomaly scores of normal instances, denoted as $\mu_\mathcal{S}$, specified by the sample mean from a known distribution $\mathcal{N}(\mu ;\sigma)$, and $\psi_K(x_j)$ diverges significantly from $\mu_\mathcal{S}$. To manage data contamination, for each training example $x_i$, we reweight the loss objectives of individual data instances with non-uniform importance weights $w_i$ determined from a constrained optimization (Sec. \ref{sec:adaptive_sample}).

\subsection{Overview of ADL}

\begin{figure*}[!ht]
\centering
  \includegraphics[width=0.8\linewidth]{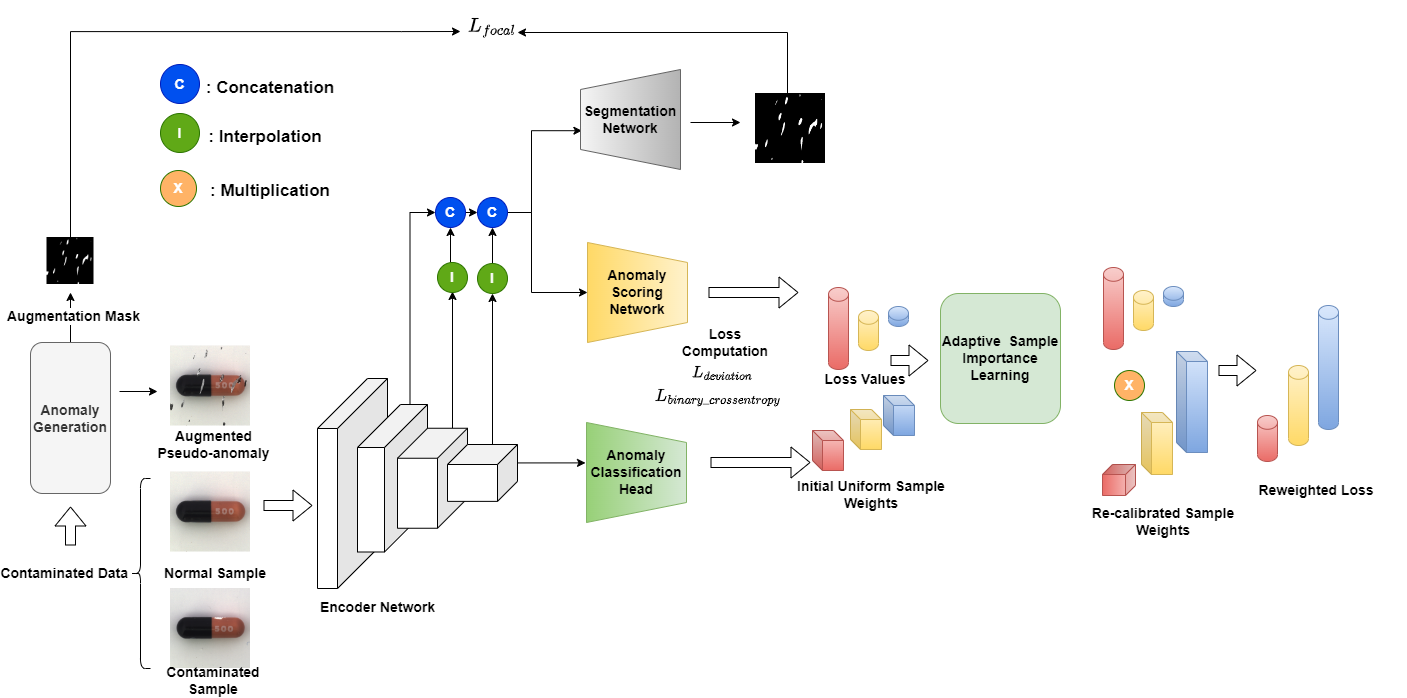}
  \caption{Overview of our proposed Adaptive Deviation Learning framework.}
  \label{fig:architec}
\end{figure*}

To address the data contamination issue, we propose a novel adaptive deviation learning framework that utilizes \textit{soft-labels} (Sec. \ref{sec:asn}) to generate anomaly scores from extracted image features, guided by a known prior distribution. Additionally, it adjusts the loss objectives of individual data instances using non-uniform importance weights, thereby effectively handling mislabeled examples.
The overall design of our proposed framework is illustrated in Fig. \ref{fig:architec}. Our framework comprises two main components: \textit{Self-supervised Robust Deviation Learning} and \textit{Adaptive Sample Importance Learning}. 

\subsection{Self-supervised Robust Deviation Learning}
The framework consists of a \textit{Synthetic Anomaly Generator}, \textit{Feature Encoder Network}, \textit{Scoring Network}, \textit{Classification head } and \textit{Segmentation Network}. The Feature Encoder extracts features from both contaminated data and pseudo-anomalies, which are then processed by the Anomaly Scoring Network and Segmentation Network, with the classification head providing soft labels to enhance the robustness of the anomaly scoring. The Segmentation Network contributes to the overall stabilization of the system.

\subsubsection{Synthetic Anomaly Generation.} 
\label{sec:synthetic}
We synthesized pseudo-anomalies using similar approach as described in \cite{zavrtanik2021draem}, with the aim of training our model to recognize anomalies based on their deviation from normality. We generate a random two-dimensional Perlin noise image \cite{perlin1985image} and then binarized it using a threshold to produce an anomaly mask $M_{p}$. An arbitrary anomaly source image $I_s$ is sampled from an unrelated external data source. The pseudo-anomaly image $I_{p}$ is generated by overlaying the anomaly source image with the anomaly mask $M_{p}$ and blending it with the normal image $I$, as follows:

\begin{equation}
I_{p} = (1 - M_{p}) \odot I +\beta (M_{p} \odot I_s) + (1 - \beta)(M_{p} \odot I) 
\label{eq:anomaly}
\end{equation}
where $\beta$ represents an opacity coefficient, randomly selected from the interval $[0.1, 1.0]$, which aids in diversifying the generated pseudo-anomalies, and $\odot$ denotes the element-wise multiplication operation. The anomaly mask generated by random Perlin noise exhibits greater irregularity and resemblance to actual anomalous shapes, in comparison to a regular rectangular mask.

\subsubsection{Feature Encoding Network.}
\label{sec:FEN}
The feature encoding network is a CNN-based network, such as ResNet \cite{he2016deep}, pretrained on ImageNet \cite{deng2009imagenet}, , which acquires feature maps of various scales and levels of abstraction (Fig. \ref{fig:architec}). As highlighted in \cite{roth2022towards},  features at higher levels of abstraction are relevant for detecting structural deviations and are useful in image-level anomaly detection. However, these features tend to be biased towards ImageNet classification and may not be suitable for anomaly detection. In contrast, mid-level features strike a balance, being neither overly generic nor biased towards specific tasks. To obtain robust features suitable for anomaly detection, the feature encoding network samples feature maps of different spatial resolutions at different depths of the CNN. These feature maps are then interpolated to have the same resolution and concatenated, following the approach outlined in \cite{defard2021padim}. Additionally, to mitigate bias and learn discriminative features on the target dataset, we employ transfer learning on the pre-trained network. Feature map extracted at layer $i$ is denoted as  $\mathcal{F}_i \in \mathbb{R}^{c_i \times H_i \times W_i }$, where $c_i$ represents number of channels, $H_i$ denotes the height, and $W_i$ stands for the width of the feature map at that layer. At the final output, we combine features into a single feature map denoted as $\mathcal{F}_{Co} \in \mathbb{R}^{C \times H \times W}$, where $C = \sum_{i=j}^{last}c_i$, $H$ is the maximum height among all layers from $j$ to the last layer, and $W$ is the maximum width among those layers. Additionally, last $\mathcal{F}_{last}$ represents the feature map from the last layer.

\subsubsection{Robust Deviation-based Anomaly Score Learning.}
\label{sec:asn}

For a given input instance $x_i$
the Anomaly Scoring Network accepts the combined feature map $\mathcal{F}_{Co}(x_i)$ as input and transforms it into anomaly score vector $\psi(x_i)= (\psi(x_{i1}), \psi(x_{i2}),..., \psi(x_{i\tilde{N}}))$ of length $\tilde{N}$. Then following the recent work on Deviation Networks \cite{pang2019deep, pang2021explainable}, we select top-$K$ $P(x_i)$ image patches of $x_i$ with highest anomaly scores. We compute the top-K anomaly score for the input image instance $x_i$ as follows:

\begin{equation}
\psi_K(x_i) = \frac{1}{K}\sum_{x_{ij} \in P(x_i)} \psi(x_{ij})  ,   
\quad \text{where} \quad |P(x_i)| = K
\label{eq:anomaly_score}
\end{equation}

After acquiring the anomaly score, we determine a reference score $\mu_\mathcal{S} \in \mathbb{R}$ is obtained according to Eqn. \ref{eq:ref_score} as outlined in previous works \cite{pang2019deep, pang2021explainable}. Here, $s_i \sim \mathcal{N}(\mu, \sigma)$  represents a randomly sampled reference score drawn from a prior Gaussian distribution, facilitating the optimization process, as the Gaussian distribution effectively models anomaly scores across diverse datasets \cite{kriegel2011interpreting}.

\begin{equation}
    \mu_{\mathcal{S}}=\frac{1}{m}\sum^{m}_{j=1}s_i
    \label{eq:ref_score}
\end{equation}

Deviation loss \cite{pang2019deep, pang2021explainable} is a contrastive loss as given by Eqn. \ref{eqn:dev} , where deviation is defined in terms of standard score.

\begin{align}
    l_{dev}(x_i) &= (1-y_i)|Z_{std}(x_i)| + y_i\max(0, \gamma - Z_{std}(x_i)), \nonumber \\
    &\quad \text{where} \quad Z_{std}(x_i) = \frac{\psi_K(x_i) - \mu_{\mathcal{S}}}{\sigma_{\mathcal{S}}} \label{eqn:dev}
\end{align}

\noindent where $\mu_{\mathcal{S}}$ , $\sigma_{\mathcal{S}}$ is sample mean and standard deviation of the set of reference scores 
$\mathcal{S}=\{s_j : j=1 \text{ to } m, \text{ and } s_j \sim \mathcal{N}(\mu, \sigma)\}$. The hyperparameter $\gamma$ is similar to the confidence interval parameter in Z-scores. The optimization process seeks to align the anomaly scores of normal examples as closely as possible with $\mu_{\mathcal{S}}$, while also guaranteeing a minimum deviation of $\gamma$ between $\mu_{\mathcal{S}}$ and the anomaly scores of anomalous samples.

However, since our data samples are corrupted, with the $\epsilon$ fraction of the normal data actually being outliers or anomalies, the hard deviation loss outlined in Eqn. \ref{eqn:dev} will result in ineffective optimization of the anomaly scoring network. To address data contamination, we've introduced the \textit{soft-deviation loss} outlined in Eqn. \ref{eqn:softdev}. This approach substitutes hard labels $y_i$ with soft labels, like estimated probabilities $p(x_i)$ from the \textit{anomaly classification head}. The network then optimizes for \textit{soft-deviation}, enabling it to learn the deviation between normality and anomalousness in the presence of data contamination.

\begin{align}
    l_{soft}(x_i) &= (1 - p(x_i)) |Z_{std}(x_i)| \nonumber \\
    &\quad + p(x_i) \max(0, \gamma - Z_{std}(x_i)) 
    \label{eqn:softdev}
\end{align}

The \textbf{anomaly classification head} accepts the feature map from the last layer, denoted as $\mathcal{F}_{last}(x_i)$, and produces an estimated probability $p(x_i)$, whether an instance $x_i$ is normal or an anomaly. The estimated probabilities are used as soft labels for our proposed \textit{soft-deviation} objective (Eqn. \ref{eqn:softdev}). The head is trained on the dataset with contaminated normal ($y_{i}=0$) and synthetic anomaly ($y_{i}=1$) samples. To minimize the impact of overfitting the probability estimation on contaminated labels during training phase, we determine intermediate levels ($\tilde{y}_{i} \in \{0,1\}$) by deploying the k-means clustering algorithm \cite{macqueen1967some} on the anomaly score (Eqn. \ref{eq:anomaly_score}). $y_{i}$ and $\tilde{y}_{i}$ are alternately used as ground truth level ($\hat{y}_i$) in the binary crossentropy loss (BCE) $l_{bce}$. The training objective aims to minimize the loss as follows:


\begin{multline}
    l_{bce}(x_i)= -(1-\hat{y}_i)\log(1-p(x_i)) -\hat{y}_i\log(p(x_i)), \\
    \text{where} \quad \hat{y}_i \in \{y_{i}, \tilde{y}_{i}\}
    \label{eq:bce}
\end{multline}

\subsection{Segmentation Network}
The anomaly segmentation network accepts the combined feature map $\mathcal{F}_{Co}(x_i)$ and generates the output segmentation mask $M_o(x_i)$. To increase the robustness and stability of the overall network, a segmentation loss $l_{seg}(x_i)$, defined by the Focal Loss \cite{lin2017focal}, is computed between $M_o(x_i)$ and the synthetic anomaly mask ground truth $M_p(x_i)$.

\subsection{Adaptive Sample Importance Learning}
\label{sec:adaptive_sample}
We utilize instance reweighting to dynamically assign relative importance weights in our robust deviation learning objective. This approach aims to mitigate the influence of contaminated data, as mislabeled instances generally exhibit higher loss values than clean samples.

Let $L(x_i, \theta)$ ($L(x_i)$ for simplicity of notation, $\theta$ is learnable parameter)  be the loss for $i$-th sample, as the training data is contaminated (associated $y_i$ is not the true class). In our deviation learning setup, this can be considered as label noise, by following \cite{kumar2021constrained}, we propose to reweight the training samples by assigning $w_i>0$, to each instance $x_i$ to optimize the following constrained optimization problem:

\begin{align}
   \min_{\theta,w:w \ge 0,\sum_{i}w_i=1}\sum_iw_iL(x_i),  \quad : Div(w,u) \leq \delta
   \label{eq:optimize}
\end{align}

\noindent Where $Div$ represents a chosen divergence metric, regulating the deviation of importance weights ($w_i$) from a reference distribution (e.g., uniform). Here, $u$ denotes the uniform distribution. The optimization is conducted over each minibatch to obviate the need for storing the entire training set, thus reducing memory overhead.

For sample reweighting, we utilize the closed-form update results as outlined in \cite{kumar2021constrained} to adjust the weights of training instances in scenarios involving KL-divergence, Reverse-KL divergence, and $\alpha$-divergence, as follows:
\paragraph{KL Divergence} is given by $Div(w,u)=KL(w,u)=\sum_iw_i\log(w_i/u_i)$, the weight $w_i$, are given by:

\begin{align}
  w_i= \frac{exp(-\frac{L(x_i)}{\lambda})} {\sum_jexp(-\frac{L(x_j)}{\lambda})}
  \label{eq:kl}
\end{align}
The Langrange multiplier $\lambda$ is a tunable hyperparameter.

\paragraph{Reverse-KL Divergence} is given by $Div(w,u)=KL(u,w)=\sum_iu_i\log(u_i/w_i)$, the weight $w_i$, are given by:

\begin{align}
  w_i= \frac{1/(L(x_i) + \lambda)}{\sum_j 1/(L(x_j) + \lambda)}
  \label{eq:rkl}
\end{align}

\paragraph{$\alpha$-divergence}
For various values of $\alpha$ ($\alpha \in \mathbb{R}$), several well-known divergences are recovered, including Reverse-KL when $\alpha=0$ and KL when $\alpha=1$. For $\alpha \in \mathbb{R} \setminus \{0,1\}$,  the weights are determined as follows:
\begin{align}
  w_i= \frac{[(1-\alpha)L(x_i) +\lambda]_{+}^{1/(\alpha-1)}} {\sum_j[(1-\alpha)L(x_j) +\lambda]_{+}^{1/(\alpha-1)}}, \quad  \alpha \neq 1
  \label{eq:genkl}
\end{align}
\noindent where $[.]_+ =max(., 0)$

By reweighing the learning objectives in Eqn.s \ref{eq:bce} and \ref{eqn:softdev} according to Eqn. \ref{eq:optimize}, the comprehensive objective aimed at addressing data contamination is expressed as follows:

\begin{align}
\min_{\theta,w_{1}, w_{2}} & \sum_i w_{1i} l_{\text{soft}}(x_i) + w_{2i} l_{\text{bce}}(x_i) +l_{seg}(x_i), \nonumber \\
\text{subject to:} & \quad \text{Div}(w_1,u) \leq \delta, \quad \text{Div}(w_2,u) \leq \delta, \nonumber \\
& \quad w_1 > 0, \quad w_2 > 0, \quad \sum_j w_{1j} = 1, \quad \sum_k w_{2k} = 1
\label{eq:overall_obj}
\end{align}
where $w_{1i}$ and $w_{2i}$ are re-calibrated sample weights for $l_{\text{soft}}$ and $l_{\text{bce}}$, respectively.

\begin{algorithm}[!t]
\caption{ADL Algorithm: Training with Contaminated Data}
\begin{flushleft}

\textbf{Input          :}
 $\mathcal{X_N}$: set of contaminated inlier training samples, contamination ratio $\epsilon$ is unknown, i.e., (1-$\epsilon$) fraction of $\mathcal{X_N}$ are true inliers and  $\epsilon$ fraction are true outliers,  $\mathcal{X_P}$:set of generated pseudo-anomalies, combined training sample set $\mathcal{X} = \mathcal{X_N} \cup \mathcal{X_P}$.

 \textbf{Output         :} Anomaly score $\psi_K: \mathcal{X} \rightarrow \mathbb{R}$.

 \textbf{Hyperparameters:} $\gamma$ for control interval in Eqn. \ref{eqn:softdev}, $\alpha$ for $\alpha$-divergence, and $\lambda$ in Eqn. \ref{eq:kl} \ref{eq:rkl}, \ref{eq:genkl}, burn-in parameter $burn\_in$
 
\end{flushleft}

\begin{algorithmic}[1]
    \STATE Randomly initialize model parameters in  $\Theta$
    \FOR{$e = 1$ to $E$} 
        \FOR{$b=1$  to $num\_of\_batches$}
        
            \STATE Sample a minibatch $\{(x_i,y_i)\}_{i=1}^{batch\_size}$

            \STATE Sample a set of $m$ reference anomaly scores randomly from $\mathcal{N}(0, 1)$. Compute the mean $\mu_{\mathcal{S}}$ and standard deviation $\sigma_{\mathcal{S}}$.
        
            \STATE for all instance in minibatch, compute soft-deviation $l_{\text{soft}}(x_i)$, binary cross-entropy $l_{\text{bce}}(x_i)$, and segmentation loss $l_{\text{seg}}(x_i)$
            \IF{$e > burn\_in$}
                \STATE Compute instance weights $w_i$ using Eqn. \ref{eq:kl}, \ref{eq:rkl} or \ref{eq:genkl} for both $l_{\text{soft}}(x_i)$ and $l_{\text{bce}}(x_i)$.
                
            \ELSE
                \STATE $w_i = \frac{1}{batch\_size} \;    \forall i$
            \ENDIF
            \STATE Compute reweighted loss $\sum_i w_{1i} l_{\text{soft}}(x_i) + w_{2i} l_{\text{bce}}(x_i) +  l_{\text{seg}}(x_i)$ as per Eqn. \ref{eq:overall_obj}
            \STATE Update model parameters in $\Theta$ using gradient of the reweighted loss, with Adam optimizer
        \ENDFOR
    \ENDFOR
\end{algorithmic}
\end{algorithm}



\section{Experiment and Analysis}
\subsection{Experimental Settings}
\textbf{Dataset.} To assess our proposed method, we have evaluated it on two publicly available benchmark datasets: the MVTec Anomaly Detection dataset \cite{bergmann2019mvtec} and the VisA dataset \cite{zou2022spot}. MVTec AD consists of 15 sub-datasets of object and texture categories, encompassing a total of 5354 images, with 1725 images designated for testing. Each sub-dataset is split into training data that includes only normal samples and test sets that contain both normal and anomalous samples. VisA comprises 12 distinct categories. It includes 8,659 normal images for training, and the testing dataset consists of 962 normal images and 1,200 anomalous images, each accompanied by ground truth pixel-level masks.

In our experimental setup, the training dataset is corrupted with anomalies, i.e., $\epsilon$ fraction (contamination ratio) of the available training data is anomalies, disguised as normal data. Since both the MVTec and VisA datasets lack anomalies in their training sets, we introduced artificial anomalies by sampling from the test set anomalies and adding zero-mean Gaussian noise with a relatively large variance to the samples. This approach is similar to LOE \cite{qiu2022latent}, where contamination occurs in the image latent space, but in our settings, we introduce contamination directly in the image space itself. For fair comparison, we applied the same strategy for anomaly contamination to all other baseline methods, except for LOE, for which we retained the original contamination strategy.
\\

\noindent \textbf{Implementation Details.}
We followed the process illustrated in Figure \ref{fig:architec} to train the proposed network. In accordance with \cite{defard2021padim, yi2020patch}, the images were resized and center-cropped to 256×256 and 224×224, respectively. The network utilizes a pre-trained ResNet-18 \cite{he2016deep} as the backbone network to extract feature maps from the original images. Feature maps are obtained from intermediate layers {Layer2, Layer3, Layer4}, as discussed in Sec. \ref{sec:FEN}. The extracted features serve as inputs to the anomaly classification head, anomaly scoring network, and segmentation network. The entire network is trained with a contaminated training set to optimize the objective described in Eqn. \ref{eq:overall_obj}. We use the Describable Textures Dataset \cite{cimpoi2014describing} as external data source for synthetic anomaly generation (Sec. \ref{sec:synthetic}). We train our model for 25 epochs on both MVTec and Visa datasets using the Adam optimizer with a learning rate of $2\mathrm{e}{-4}$. We set $K$=$0.1$ in top-K anomaly score computation (Eqn. \ref{eq:anomaly_score}). For prior reference score we choose Standard Normal distribution $\mathcal{N}(0,1)$ with number of reference samples $m$ is set to 5000 (Eqn. \ref{eq:ref_score}). We choose a confidence interval of $\gamma$ = $5$ (Eqn. \ref{eqn:dev}). The divergence parameter $\alpha$ is set to 0.1 and the lagrangian multiplier $\lambda$ is fixed to 0.1 (Eqn. \ref{eq:kl}, \ref{eq:rkl}, \ref{eq:genkl}).
\\

\noindent \textbf{Baselines and Metrics.}
We evaluate our proposed method against leading state-of-the-art techniques for anomaly detection, specifically PatchCore \cite{roth2022towards}\footnote[2]{PatchCore, DestSeg and DR\AE M are taken from official implementations in pytorch : https://github.com/amazon-science/patchcore-inspection, https://github.com/apple/ml-destseg and https://https://github.com/VitjanZ/DRAEM/ respectively}, DestSeg \footnotemark[2] \cite{zhang2023destseg}, DR\AE M \cite{zavrtanik2021draem} \footnotemark[2], and LOE \cite{qiu2022latent} \footnote[3]{We have used authors released code for \textit{soft}-LOE https://github.com/boschresearch/LatentOE-AD which uses ResNet152 as backbone}. Since PatchCore and DR\AE M are originally designed to work exclusively with clean normal samples, we modified their code to include data contamination in the training set, as used in our approach, to ensure a fair comparison.


In our experiments, we employ standard performance metrics for detection: the Image-level Area Under the Receiver Operating Characteristic Curve (AUC-ROC) and the Area Under the Precision-Recall Curve (AUC-PR). The AUC-ROC metric illustrates the relationship between true positives and false positives on the ROC curve. An AUC-ROC value of one implies the best performance. The AUC-PR metric summarizes the Precision versus Recall curve and is especially useful for imbalanced datasets where negative examples significantly outnumber the positive ones. An AUC-PR value of one indicates perfect precision and recall, while a value near 0.5 for either AUC-ROC or AUC-PR denotes performance equivalent to random guessing.

\subsection{Results and Analysis}

In this section, we will evaluate our model's performance against other competitive AD methods and analyze our proposed approach from different viewpoints to achieve a through understanding of its strengths and effectiveness.
\\

\begin{figure}[!ht]
  \centering
    \includegraphics[width=0.8\linewidth]{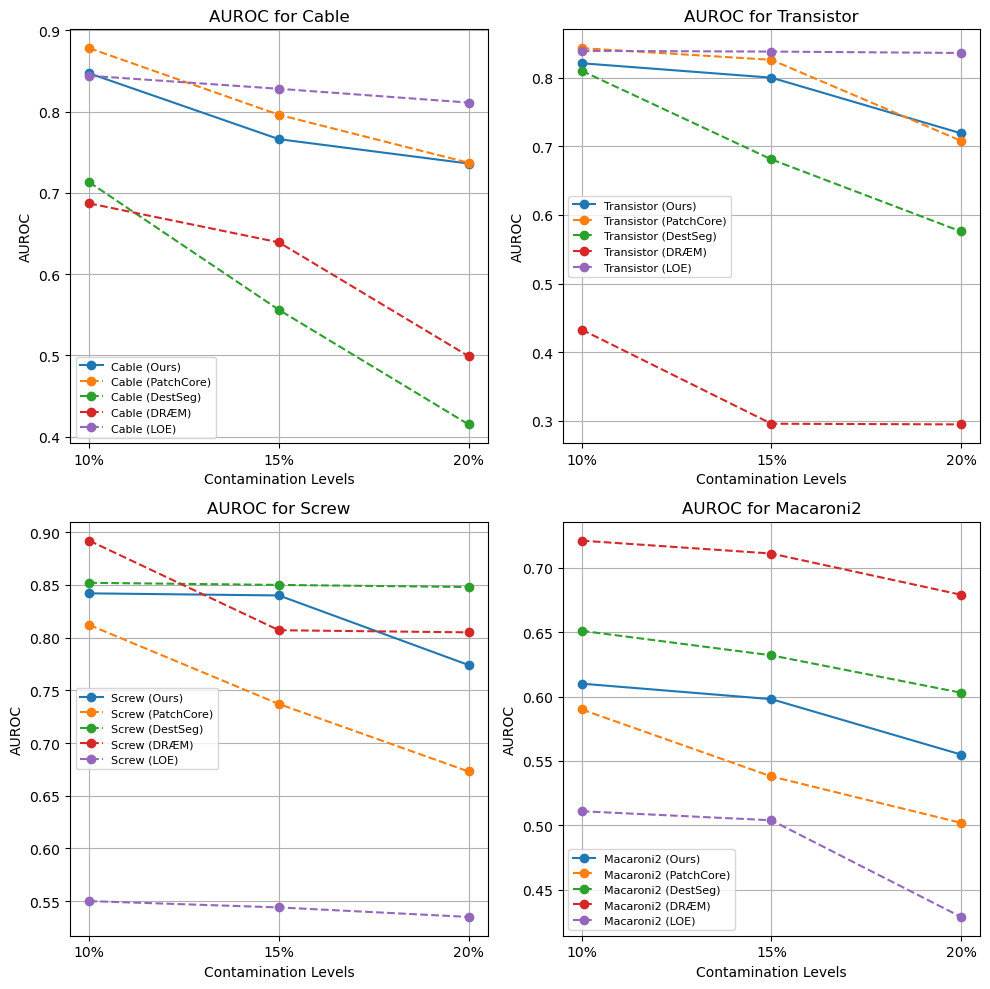}
    \caption{Key Failure Cases: Sensitivity to Contamination Levels (10\%-20\%) on MVTec and VisA Datasets.}
    \label{fig:failuere}
  
\end{figure}

\noindent \textbf{Main Results.} The results of our proposed method, along with those of prior state-of-the-art approaches across all datasets, are presented for various contamination ratios ($\epsilon \%$) in Table \ref{tab:image_roc}. The AUC-ROC values indicate that our method outperforms the top-performing DestSeg model at contamination levels of 15\% and 20\%, and surpasses other competitive baselines across all contamination levels. For the VisA dataset, our proposed approach exceeds the performance of all its competitors. These results clearly highlight the effectiveness of our proposed method. As depicted in Figure \ref{fig:failuere}, we have identified failure cases in which classes such as cable, transistor, screw, and macaroni2 showed significant performance drops (9-13\%). However, in most instances, LOE maintained stable performance across contamination levels, while DestSeg demonstrated greater stability for macaroni2.

\begin{table*}[ht]

\centering
\caption{Image-level AUC-ROC \% results for anomaly detection on the MVTec AD and VisA datasets with various contamination levels (10\%, 15\%, 20\%). The values represent the mean results from multiple runs. The highest results are highlighted in bold, while the second highest results are underlined.}
\scalebox{0.7}{
\begin{tabular}{llllll} 
\toprule
Model &PatchCore & DestSeg &  DR{\AE}M & LOE & Ours \\
\midrule
 MVTec Classes &10\%/15\%/20\% &10\%/15\%/20\% &10\%/15\%/20\% &10\%/15\%/20\% 
&10\%/15\%/20\% \\
\midrule
 carpet  &0.775/0.675/0.662 &\underline{0.938}/0.832/0.76 &0.802/0.763/0.729 &0.937/\underline{0.926}/\textbf{0.921} &\textbf{0.943}/\textbf{0.93}/\underline{0.886}
 \\
  bottle & 0.888/0.827/0.757 &\textbf{1}/\textbf{1}/\textbf{1} &0.963/0.959/0.952 & \underline{0.98}/\underline{0.98}/0.978 &\textbf{1}/\textbf{1}/\underline{0.999}
 \\
  cable &	\textbf{0.878}/\underline{0.796}/\underline{0.737}&	0.713/0.556/0.415 &	0.687/0.639/0.499 &0.844/\textbf{0.828}/\textbf{0.811} &\underline{0.847}/0.766/0.736
 \\
   capsule &0.857/0.784/0.781 &\underline{0.941}/\textbf{0.941}/\textbf{0.94} &\textbf{0.947}/\underline{0.892}/\underline{0.891} &0.791/0.781/0.767 &0.862/0.858/0.823
 \\
   zipper &0.871/0.788/0.793 &\textbf{0.999}/\textbf{0.999}/\textbf{0.999} &\textbf{0.999}/\underline{0.99}/\underline{0.99} &0.879/0.878/0.87 &\underline{0.96}/0.954/0.953
 \\
   wood &0.733/0.631/0.569 &0.918/\underline{0.916}/0.892 &0.711/0.641/0.64 &\underline{0.935}/0.912/\underline{0.898} &\textbf{0.998}/\textbf{0.998}/\textbf{0.997}
 \\
   transistor &\textbf{0.843}/\underline{0.826}/0.708 &0.81/0.681/0.576 &0.433/0.296/0.295 &\underline{0.839}/\textbf{0.838}/\textbf{0.836} &0.821/0.8/\underline{0.719}
 \\
   toothbrush &0.864/0.745/0.733 &\underline{0.889}/0.884/\underline{0.808} &\textbf{0.91}/\textbf{0.903}/0.794 &\textbf{0.91}/\underline{0.9}/\textbf{0.892} &0.853/0.821/0.764
 \\
   tile &0.836/0.756/0.742 &\underline{0.96}/\underline{0.939}/0.869 &0.941/0.931/0.806 &\textbf{0.993}/\textbf{0.992}/\textbf{0.991} &0.952/0.924/\underline{0.923}
 \\
   screw &0.812/0.737/0.673 &\underline{0.852}/\textbf{0.85}/\textbf{0.848} &\textbf{0.892}/0.807/\underline{0.805} &0.55/0.544/0.535 &0.842/\underline{0.84}/0.774
 \\
   pill &0.804/0.741/0.661 &\textbf{0.928}/\textbf{0.92}/\textbf{0.916} &0.894/0.876/\underline{0.875} &0.751/0.733/0.712 &\underline{0.916}/\underline{0.885}/0.872
 \\
   metal\_nut &0.846/0.752/0.68 &0.972/0.915/\underline{0.86} &\underline{0.975}/\underline{0.942}/\textbf{0.895} &0.735/0.71/0.71 &\textbf{0.981}/\textbf{0.973}/\underline{0.86}
 \\
   leather &0.79/0.712/0.689 &\textbf{1}/\textbf{1}/\textbf{1} &\textbf{1}/\textbf{1}/\textbf{1} &0.989/0.985/0.971 &\underline{0.999}/\underline{0.999}/\underline{0.988}
 \\
   hazelnut &0.637/0.616/0.651 &\textbf{0.993}/\textbf{0.981}/\textbf{0.978} &0.979/0.96/0.878 &0.958/0.952/0.942 &\underline{0.984}/\underline{0.979}/\underline{0.971}
 \\
   grid &0.624/0.601/0.515 &\textbf{0.999}/\textbf{0.996}/\textbf{0.995} &\underline{0.98}/\underline{0.918}/0.796 &0.473/0.436/0.4 &0.895/0.887/\underline{0.827}
 \\
   \textbf{Average} &0.804/0.732/0.69 &\textbf{0.927}/\underline{0.894}/\underline{0.857} &0.879/0.834/0.79 &0.838/0.826/0.816 &\underline{0.924}/\textbf{0.908}/\textbf{0.873}
 \\
\midrule
 VisA Classes &10\%/15\%/20\% &10\%/15\%/20\% &10\%/15\%/20\% &10\%/15\%/20\% 
&10\%/15\%/20\% \\
\midrule
 capsules &0.548/0.489/0.484  &\underline{0.66}/\underline{0.63}/\underline{0.627} &0.462/0.442/0.413 
 &0.622/0.578/0.535   &\textbf{0.751}/\textbf{0.744}/\textbf{0.724}
 \\
  candle &0.751/0.742/0.737 &\textbf{0.918}/\textbf{0.91}/\textbf{0.9}	 &0.822/0.734/0.662 &0.864/0.843/0.81 &\underline{0.911}/\underline{0.906}/\underline{0.878}
 \\
  cashew &0.771/0.754/0.727 &0.849/0.838/0.827 &0.432/0.309/0.24 &\textbf{0.944}/\textbf{0.928}/\textbf{0.918} &\underline{0.908}/\underline{0.898}/\underline{0.869}
 \\
   chewinggum &0.714/0.582/0.543		&0.85/0.835/0.83		&0.807/0.754/0.712		&\textbf{0.959}/\textbf{0.958}/\textbf{0.95}		&\underline{0.927}/\underline{0.92}/\underline{0.914}
 \\
   macaroni1 &0.689/0.683/0.664		&0.745/0.735/\underline{0.729}		&\textbf{0.862}/\textbf{0.849}/\textbf{0.824}		&0.67/0.656/0.598		&\underline{0.806}/\underline{0.738}/\underline{0.729}
\\
   macaroni2 &0.59/0.538/0.502		&\underline{0.651}/\underline{0.632}/\underline{0.603}		&\textbf{0.721}/\textbf{0.711}/\textbf{0.679}	&0.511/0.504/0.429		&0.61/0.598/0.555
 \\
   fryum &0.77/0.713/0.696		&\underline{0.834}/\underline{0.829}/\underline{0.8} &0.751/0.664/0.556 &0.754/0.746/0.735 &\textbf{0.931}/\textbf{0.925}/\textbf{0.912}
\\
   pipe\_fryum &0.761/0.629/0.595 &\underline{0.926}/\underline{0.91}/\underline{0.9} &\textbf{0.955}/\textbf{0.947}/\textbf{0.944} &0.81/0.808/0.801 &0.915/0.908/0.879
 \\
   pcb1 &0.734/0.728/0.724		&\textbf{0.85}/\textbf{0.842}/\textbf{0.83}	&0.21/0.142/0.114	&0.816/0.794/0.777	&\underline{0.82}/\underline{0.817}/\underline{0.784}
 \\
   pcb2 &0.716/0.707/0.697		&0.789/0.767/\underline{0.758}	&0.232/0.196/0.131		&\textbf{0.847}/\underline{0.785}/0.741	&\underline{0.822}/\textbf{0.802}/\textbf{0.789}
\\
   pcb3 &0.663/0.654/0.645		&\textbf{0.924}/\textbf{0.898}/\textbf{0.868}		&0.704/0.603/0.452		&0.797/0.756/0.711		&\underline{0.812}/\underline{0.809}/\underline{0.791}
 \\
   pcb4 &0.889/0.879/0.871		&\underline{0.963}/\textbf{0.962}/\textbf{0.941}		&0.805/0.756/0.777		&\textbf{0.966}/\underline{0.93}/\underline{0.909}	&0.91/0.864/0.858
 \\
   \textbf{Average} &0.716/0.675/0.657		&\underline{0.83}/\underline{0.816}/\underline{0.801}	&0.647/0.592/0.542 &0.797/0.774/0.743 &\textbf{0.844}/\textbf{0.827}/\textbf{0.807}
 \\

\bottomrule
\label{tab:image_roc}
\end{tabular}
}
\end{table*}

\begin{figure*}[!ht]
  \centering
    \includegraphics[width=0.8\linewidth]{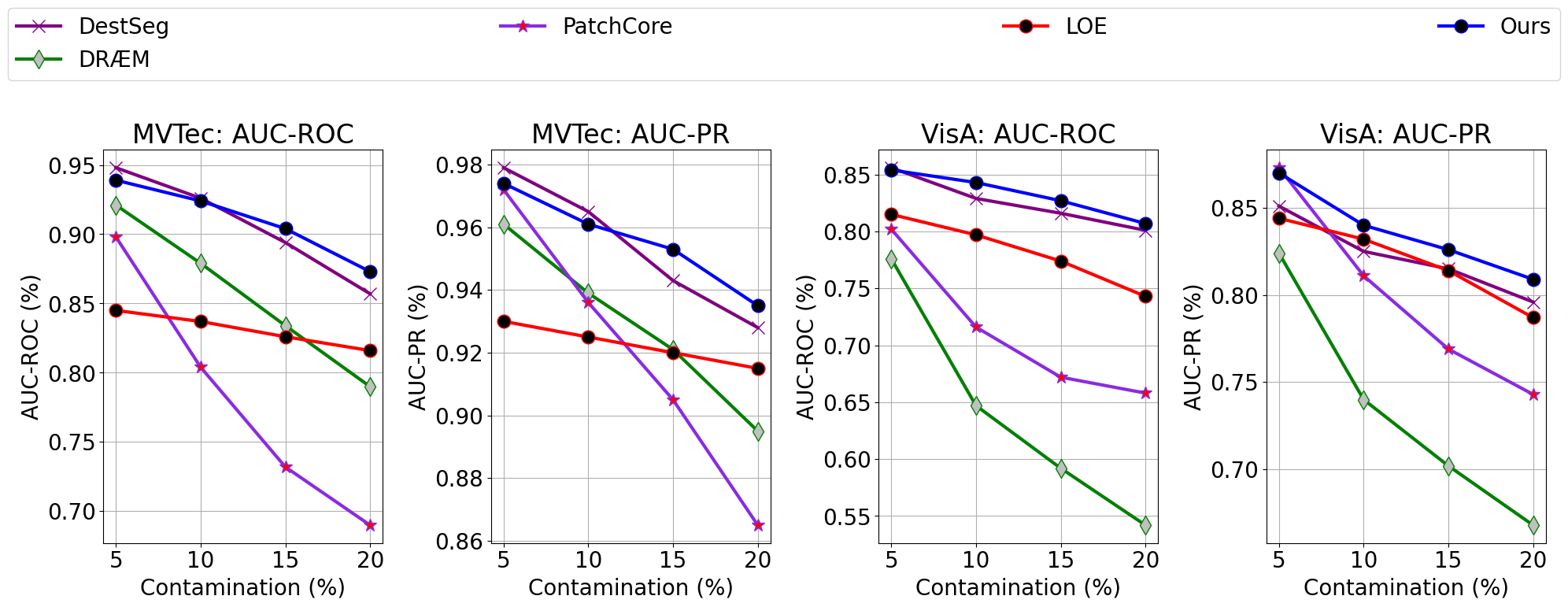}
    \caption{Robustness: AUC-ROC and AUC-PR vs. Anomaly Contamination (\%) on MVTec and VisA datasets.}
    \label{fig:robust_roc}
  
\end{figure*}

\noindent{\textbf{Robustness.}} We investigate the robustness of our proposed framework and examine how increased contamination levels in the training data impact performance on evaluation data for both the MVTec and VisA datasets, considering both average AUC-ROC and AUC-PR metrics. We varied the contamination ratio from 5\% to 20\%. We compare the performance of our model with that of other baselines. As depicted in Figure \ref{fig:robust_roc}, all methods exhibit decreased performance as contamination ratio rises. However, methods such as PatchCore and DR\AE M experience notably sharp decreases with higher contamination levels. In contrast, approaches like LOE and DestSeg demonstrate greater resilience. Figure \ref{fig:robust_roc} illustrates that DestSeg performs well at lower contamination levels. However, as contamination levels increase, our method demonstrates improved resilience and overall superior performance compared to other state-of-the-art methods.

\begin{figure}[!ht]
  \centering
  \
    \begin{subfigure}[b]{0.49\linewidth}
    \includegraphics[width=1.0\linewidth]{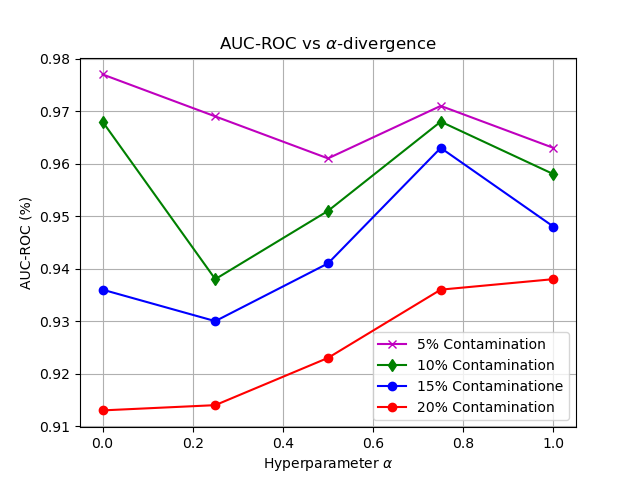}
    \caption{Sensitivity to $\alpha$-divergence at different contamination levels }
    
    \label{fig:hyp_alpha}
    \end{subfigure}
  \hfill
    \begin{subfigure}[b]{0.49\linewidth}
    \includegraphics[width=1.0\linewidth]{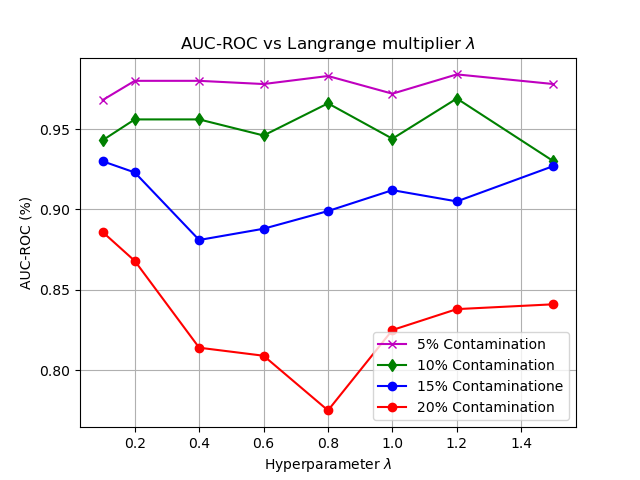}
    \caption{Sensitivity to Lagrange Multiplier $\lambda$ at different contamination levels }
    
    \label{fig:hyp_lambda}
    \end{subfigure}
\caption{Sensitivity: AUC-ROC vs Hyperparameters at different levels of contamination (\%) on MVTec dataset.}
\label{fig:hyp}
\end{figure}

\noindent \textbf{Sensitivity.}
In order to understand the impact of the Lagrange Multiplier $\lambda$ and $\alpha$ (Eqn.s \ref{eq:kl}, \ref{eq:rkl}, and \ref{eq:genkl}), we varied $\lambda$ from 0.1 to 1.5 and $\alpha$ from 0 to 1 respectively. We studied the sensitivity of the proposed approach at different contamination levels, as shown in Fig. \ref{fig:hyp}.
The model demonstrates reasonable resilience against variations in both hyperparameters. Regarding the Lagrange multiplier $\lambda$, we observed a dip in AUC-ROC values at higher contamination levels (20\%), occurring approximately between 0.65 and 0.9. The model's performance remains stable within the range of 1 to 1.5.

\begin{table}[ht]

\centering
\caption{Ablation Study: Average AUC-ROC (\%) on MVTec and VisA datasets on 10\% contamination level.}
\scalebox{0.6}{
\begin{tabular}{llllllll} 
\toprule
  Module &DL & Wt-DL & DL-CE &SoftDL-CE & Wt-SoftDL-CE & Proposed \\
\midrule
  $l_{dev}$ &\checkmark&\checkmark&\checkmark&&& \\
  $l_{bce}$ &&&\checkmark&\checkmark&\checkmark&\checkmark \\
  $l_{soft}$&&&&\checkmark&\checkmark&\checkmark \\
  \textit{Sample reweight}&&\checkmark&&&\checkmark&\checkmark \\
  $l_{seg}$&&&&&&\checkmark \\
\midrule  

MVTec (Average) &0.842&0.851&0.865&0.881&0.892&\textbf{0.924} \\
 
VisA (Average) &0.788&0.791&0.795&0.805&0.821&\textbf{0.843} \\

\bottomrule
\label{tab:ablation}
\end{tabular}
}
\end{table}

\noindent \textbf{Ablation Study.} We investigate the impact of loss objectives in our proposed framework and introduce five variants for the ablation study: i) \textbf{\textit{DL:}} This variant uses solely deviation loss as the objective function. ii)\textbf{\textit{Wt-DL:}} Here, we apply sample reweighted deviation loss. iii) \textbf{\textit{DL-CE:}} This variant combines standard deviation loss with BCE loss as the objective function. iv) \textbf{\textit{SoftDL-CE:}} In this case, we replace the deviation loss with the proposed soft deviation loss. v)\textbf{\textit{Wt-SoftDL-CE:}} This variant uses the proposed reweighted soft deviation loss as the objective function.

Table \ref{tab:ablation}  summarizes the results of the ablation study conducted with fixed contamination settings (10\%). Replacing the deviation loss with soft deviation loss improves overall model performance on both datasets. While the inclusion of sample reweighting leads to a slight improvement, it falls short of the desired impact. However, combining both sample reweighting and soft deviation, as proposed in adaptive deviation learning, results in further performance gains. The likelihood-based soft-deviation approach reduces the impact of contaminated samples by ensuring the anomaly scores of clean nominal samples stay near the reference mean, using an auxiliary classification head for likelihood prediction. In contrast, adaptive sample reweighting mitigates contamination effects by directly adjusting the loss value. Additionally, our experiments demonstrate that integrating segmentation loss into our model objective not only enhances stability but also improves average performance.

\section{Conclusion and Future Work}
In this paper, we introduce an Adaptive Deviation Learning framework for detecting visual anomalies, which incorporates dynamic instance reweighting and a likelihood-based soft deviation objective to compute anomaly scores in the presence of data contamination. The framework approximates anomaly scores of normal samples using a prior distribution and employs a soft deviation loss based on standard scores to drive outlier anomaly scores further from the reference mean. Comprehensive evaluation on benchmark MVTec and VisA demonstrates that our model outperforms existing methods significantly under substantial data contamination. In future work, we plan to investigate the efficacy of our method in anomaly localization and explore its explainability to enhance transparency and understanding of the decision-making process for end-users.

\subsection*{Acknowledgements}
This work was partially supported by the Wallenberg AI, Autonomous Systems and Software Program (WASP) funded by Knut and Alice Wallenberg Foundation. 

{\small
\bibliographystyle{ieee_fullname}
\bibliography{egbib}
}

\end{document}